# Inferring knowledge from a large semantic network

Dominique Dutoit

Memodata et CRISCO
17, rue Dumont d'Urville
F-14000 Caen

memodata@wanadoo.fr

Thierry Poibeau

Thales and LIPN
Domaine de Corbeville
F-91404 Orsay

thierry.poibeau@thalesgroup.com

## Abstract

In this paper, we present a rich semantic network based on a differential analysis. We then detail implemented measures that take into account common and differential features between words. In a last section, we describe some industrial applications.

## 1 Introduction: textual and differential semantics

In textual analysis, each lexical item from a text is broken down in a list of semantic features. Features are intended to differentiate one word from another: a naive example would be a feature `back` that could express the difference between a chair and a stool. Of course, most of the time, features are not so easy to define. Some feature typologies have been provided, but there are still much discussions about the nature of a feature in a text. Most of the studies concerning differential semantics are based on a human approach to texts (this can lead to different problems, see below). Textual Semantics, also called differential semantics, is revisiting the concepts of continental structuralism like decomponential semantics (Cavazza, 1998).

The problem is then to have a lexical formalism that allows, for a lexical item, a simple description and some other features which could be dynamically inferred from the text. For example, the dictionary should mention that a "door" is an aperture, but it is more questionable to mention in the dictionary that "one can walk through a door". However, it can be an important point for the interpretation of a sentence in context.

That is the reason why Pustejovsky introduced in the nineties the notion of "generative lexicon" (Pustejovsky, 1991) (Pustejovsky, 1995). His analysis has to deal with the notion of context: he proposes to associate to a word a core semantic description (the fact that a "door" is an "aperture") and to add some additional features, which can be activated in context ("walk-through" is the *telic role* of a "door"). However, Pustejovsky does not take into account important notions such as lexical chains and text coherence. He proposes an abstract model distant from real texts.

Semantic features can be used to check out text coherence through the notion of "isotopy". This notion is "the recurrence within a given text section (regardless of sentence boundaries) of the same semantic feature through different words" (Cavazza, 1998). The recurrences of these features throughout a text allows to extract the topic of interest and some other points which are marginally tackled in the text. It provides interesting ways to glance at the text without a full reading of it; it also helps the interpretation.

In this paper, we present a rich semantic network based on a differential analysis. We then detail implemented measures that take into account common and differential features between words. In a last section, we describe some industrial applications.

## 2 The semantic network

The semantic network used in this experiment is a multilingual network providing information for 5 European languages. We quickly describe the network and then give some detail about its overall structure.

## 2.1 Overall organisation

The semantic network we use is called *The Integral Dictionary*. This database is basically structured as a merging of three semantic models available for five languages. The maximal coverage is given for the French language, with 185.000 word-meanings encoded in the database. English Language appears like the second language in term of coverage with 79.000 word-meanings. Three additional languages (Spanish, Italian and German) are present for about 39.500 senses.

These smallest dictionaries, with universal identifiers to ensure the translation, define the Basic Multilingual Dictionary available from the ELRA. Grefenstette (1998) has done a corpus coverage evaluation for the Basic Multilingual Dictionary. The newspapers corpora defined by the US-government-sponsored Text Retrieval Conference (TREC, 2000) has been used as a test corpus. The result was that the chance of pulling a random noun out of the different corpus was on average 92%. This statistic is given for the Basic Multilingual Dictionary and, of course, the French Integral Dictionary reaches the highest coverage.

This semantic network is richer than Wordnet (Bagga *et al.*, 1997) (Fellbaum, 1998): it has got a larger number of links and is based on a componential lexical analysis. Because words are highly interconnected, the semantic network is easily tunable for a new corpus (see section 2.3).

## 2.2 Measure of distance between words

We propose an original way to measure the semantic proximity between two words. This measure takes into account the similarity between words (their common features) but also their differences.

Let's take the following example:

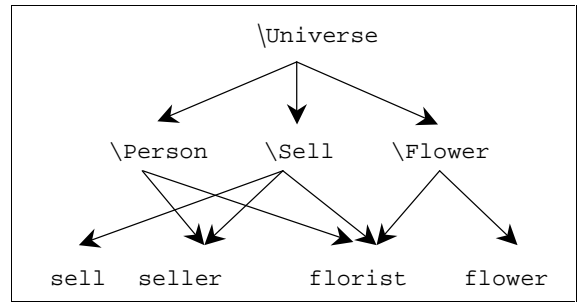

**Figure 1: An example of semantic graph**

The comparison between two words is based on the structure of the graph: the algorithm calculates a score taking into account the common ancestors but also the different ones. Let's take the example of *seller* and *florist*. They have two common ancestors: `\Person` and `\Sell`, but also one differential element: the concept `\Flower` that dominates `florist` but not `seller`.

The notion of "nearest common ancestor" is classical in graph theory. We extend this notion to distinguish between "symmetric nearest common ancestor" (direct common ancestor for both nodes) and "asymmetric nearest common ancestor" (common ancestor, indirect at least for one node).

**Definition: Distance between two nodes in a graph**

We note d the distance between two nodes A and B in a graph. This distance is equivalent to the number of intervals between two nodes A and B. We have `d(A, B) = d(B,A)`.

**Example:** We have `d(sell, \Sell) = 1` and `d(sell, \Universe) = 2`, from Figure 1. Note that `d(sell, \Sell) = d(\Sell, sell) = 1`.

Given:
   h(f) = the set of ancestors of f .
   c(f) = the set of arcs between a daughter f and the graph's root.

We have:
```
h(seller) = {\Sell, \Person, \Universe}
c(seller) = { (seller, \Sell), (seller, \Person), (\Sell, \Universe), (\Person, \Universe)}

etc.
```

**Definition: Nearest common ancestors (NCA)**

The nearest common ancestors between two words A and B are the set of nodes that are daughters of `c(A) ∩ c(B)` and that are not ancestors in `c(A) ∩ c(B)`.

**Example:** From Figure 1, we have:
```
c(seller) ∩ c(florist) = { (\Sell,
  \Universe), (\Person, \Universe) }
DaughterNodes(c(seller) ∩
  c(florist)) = { \Sell, \Person }
AncestorNodes (c(seller) ∩
  c(florist)) = { \Universe }
```

The NCA is equal to the set of nodes in the set `DaughterNodes (c(seller) ∩ c(florist))` but not in `AncestorNodes (c(seller) ∩ c(florist))`. Given that no element from AncestorNodes (c(seller) ∩ c(florist)) appears in DaughterNodes(c(seller) ∩ c(florist)), we have:
```
NCA(seller, florist) = { \Sell,
  \Person }
```

We then propose a measure to calculate the similarity between two words. The measure is called *activation* and only takes into account the common features between two nodes in the graph. An equal weight is attributed to each NCA. This weight corresponds to the minimal distance between the NCA and each of the two concerned nodes.

**Definition: activation ($d_\curlywedge$)**

The *activation* measure $d_\curlywedge$ is equal to the mean of the weight of each NCA calculated from A and B:

$$d_\curlywedge(A, B) = \frac{1}{n}\sum_{i=1}^{n}(d(A,NCA_i)+d(B,NCA_i))$$

The activation measure has the following properties:
- $d_\curlywedge$(A, A) = 0, because A is the unique NCA of A $\curlywedge$ A.
- $d_\curlywedge$(A, B) = $d_\curlywedge$(B, A) (symmetry)
- $d_\curlywedge$(A, B) + $d_\curlywedge$(B, C) >= $d_\curlywedge$(A, C) (euclidianity)

**Example :** According to Figure 1, we have NCA(seller, florist) = { \Sell, \Person}. Consequently, if we assign a weight equal to 1 to each link, we have:

```
d_\curlywedge(seller, florist) = (d(seller,
  \Sell)+d(\Sell, florist) +
  d(seller, \Person)+ d(\Person,
  florist)) / 2
d_\curlywedge(seller, florist) = (1 + 1 + 1 +
  1) / 2
d_\curlywedge(seller, florist) = 2
```

We can verify that:
```
d_\curlywedge(florist, seller) = d_\curlywedge(seller,
  florist) = 2
```

The set of NCA takes into account the common features between two nodes A et B. We then need another measure to take into account their differences. To be able to do that, we must define the notion of asymmetric nearest common ancestor.

**Definition: Asymmetric nearest common ancestor (ANCA)**

The asymmetric nearest common ancestors from a node A to a node B is contained into the set of ancestors of `c(B) ∩ c(A)` which have a direct node belonging to `h(A)` but not to `h(B)`.

**Example:** According to Figure 1, we have:
```
AncestorNodesNotNCA (c (seller) ∩
  c(florist)) = { \Universe }
```

The concept `\Universe` does not have any daughter that is a member of `h(seller)` but not of `h(florist)`. As a consequence, we have:
```
ANCA(seller, florist) = ∅
```

On the other hand, the concept `\Universe` has a daughter `\Flower` that belongs to `h(florist)` but not to `h(seller)`. As a consequence, we have:
```
ANCA(florist, seller) = {\Universe}
```

It is now possible to measure the distance between two words from their differences. A weight is allocated to each link going from node $N_i$, asymmetric nearest common ancestor, to A and B. The weight is equal to the length of the minimal length of the path going from A to $N_i$ and from B to $N_i$.

**Definition: proximity ($d_\perp$)**

The proximity measure takes into account the common features but also the differences between two elements A and B and is defined by the following function:

$$d_\perp(A,B) = d_\curlywedge(A,B) + \frac{1}{n}\sum_{i=1}^{n}(d(A, ANCA_i) + d(B, ANCA_i))$$

Because the set of ANCA from a node A to a node B is not the same as the one from a node B to a node A, the proximity measure has the following properties:

- $d_\perp(A, A) = 0$, because $ANCA(A, A) = \emptyset$.
- $d_\perp(A, B) \neq d_\perp(B, A)$ if the set of ANCA is not empty (antisymmetry)
- $d_\perp(A, B) + d_\perp(B, C) >= d_\perp(A, C)$ (euclidianity)

The proximity measure is dependent from the structure of the network. However, one must notice that this measure is a relative one: if the semantic network evolves, all the proximity measures between nodes are changed but the relations between nodes can stay relatively stable (note that the graph presented on Figure 1 is extremely simplified: the real network is largely more connected).

**Example:** Let's calculate the semantic proximity between `seller` and `florist`: $d_\perp$ (`seller, florist`). We will then be able to see that the proximity between `florist` and `seller` does not produce the same result (antisymmetry).

Given that ANCA(`seller, florist`) = $\emptyset$, the second element of the formula based on the set of ANCA is equal to 0. We then have:

```
d⊥(seller, florist) =
  d⋏(seller, florist) + 0
d⊥(seller, florist) = 2 + 0
d⊥(seller, florist) = 2
```

ANCA(`seller, florist`) is the set containing the concept \Universe, because the concept \Flower is an ancestor of `florist` but not of `seller`. We then have:

```
d⊥(florist, seller) = d⋏ (florist,
   seller) + (d(seller, \Universe) +
   d(\Universe, florist)) / 1
d⊥(florist, seller) = 2 + ( 2 + 2 ) /
   1
d⊥(florist, seller) = 6
```

To sum up, we have:

```
d⋏(florist, seller) = 2
d⋏(seller, florist) = 2
d⊥(seller, florist) = 2
d⊥(florist, seller) = 6
```

The proximity measure discriminates `florist` from `seller`, whereas the activation measure is symmetric. The componential analysis of the semantic network reflects some weak semantic differences between words.

### 2.3 Link weighting

All the links in the semantic network are typed so that a weight can be allocated to each link, given its type. This mechanism allows to very precisely adapt the network to the task: one does not use the same weighting to perform lexical acquisition as to perform word-sense disambiguation. This characteristic makes the network highly adaptive and appropriate to explore some kind of lexical tuning.

## 3 Experiment and evaluation through an information filtering task

In this section we propose to evaluate the semantic network and the measures that have been implemented through a set of NLP applications related to information filtering. To help the end-user focus on relevant information in texts, it is necessary to provide filtering tools. The idea is that the end-user defines a "profile" describing his research interests (van Rijsbergen, 1979) (Voorhees, 1999).

A profile is a set of words, describing the user's domain of interest. Unfortunately the measures we have described are only concerned with simple words, not with set of words.

We first need to slightly modify the activation measure, so that it accepts to compare two sets of words, and not only two simple words[1]. We propose to aggregate the set of nodes in the graphs corresponding to the set of words in the profile. This node has the following properties:

$$h(M) = \bigcup_{i=1}^{n} h(m_i)$$

$$c(M) = \bigcup_{i=1}^{n} c(m_i)$$

where `h(M)` is the set of ancestors of `M` and `c(M)` the set of links between `M` and the root of the graph. It is then possible to compare two set of words, and not only two simple words.

In the framework of an Information Extraction task, we want to filter texts to focus on sentences that are of possible interest for the extraction process (sentences that could allow to fill a given slot). We then need a very precise filtering process performing at the sentence level[2]. We used the activation measure for the filtering task. A sentence is kept if the activation score between the filtering profile and the sentence is above a certain threshold (empirically defined by the end-user). A filtering profile is a set of words in relation with the domain or the slot to be fill, defined by the end-user.

We made a set of experiments on a French financial newswire corpus. The topic was the same as in the MUC-6 conference (1995): companies purchasing other companies. We made the experiment on a set of 100 news stories (no training phase).

The filtering profile was composed of the following words: `rachat`, `cession`, `enterprise` (buy, purchase, company). The corpus has been manually processed to identify relevant sentences (the reference corpus). We then compare the result of the filtering task with the reference corpus.

In the different experiments we made, we modified different parameters such as the filtering threshold (the percentage of sentences to be kept). We obtained the following results:

|           | 10% | 20% | 30% | 40% | 50% |
|-----------|-----|-----|-----|-----|-----|
| Precision | .72 | .54 | .41 | .33 | .28 |
| Recall    | .43 | .64 | .75 | .81 | .85 |

We also tried to normalize the corpus, that is to say to replace entities by their type, to improve the filtering process. We used a state-of-the-art named entity recogniser that was part of a larger toolbox for named entity recognition.

|           | 10% | 20% | 30% | 40% | 50% |
|-----------|-----|-----|-----|-----|-----|
| Precision | .75 | .56 | .43 | .34 | .29 |
| Recall    | .49 | .71 | .82 | .89 | .94 |

We notice that we obtain, from 10% of the corpus, a 75% precision ratio (3 sentences out of 4 are relevant) and nearly a 50% recall ratio. The main interest of this process is to help the end-user directly focus on relevant pieces of text. This strategy is very close from the EXDISCO system developed by R. Yangarber at NYU (2000), even if the algorithms we use are different.

## 4 Application services overview

In this section, we detail some of the applications developed from the semantic network described above. All of these applications are available through java API. They are part of the applicative part of the network called the Semiograph[3]. Most of the examples will be given in French.

### 4.1 Query expansion

This application gives a help to the users who query the web through a search engine. In this framework, the Semiograph has to determinate the sense of the query and generate (or suggest) an expansion of the query in accordance to the semantic and syntactic properties of the source.

---

[1] This measure allows to compare two set of words, or two sentences. For a sentence, it is first necessary to delete empty words, to obtain a set of full words

[2] This is original since most of the systems so far are concern with texts filtering, not sentence filtering.

[3] Part of Speech tagging, syntactic analysis for French and Word Sense Disambiguation are also APIs of the Semiograph.

The Semiograph links independent mechanisms of expansion defined by the user. Eight mechanisms are available :

- Alias: to get the graphics variant
- Synonyms: to get synonyms for a meaning
- Hypernyms: to get hypernyms for a meaning
- Hyponyms: to get hyponyms for a meaning
- Inflected forms : to get the inflected for a meaning
- Derived forms: to get correct lexical functions in accordance or not with the syntactical proposition
- Geographical belonging: to get toponyms
- Translation (language parameter) : to get a translation of the query.

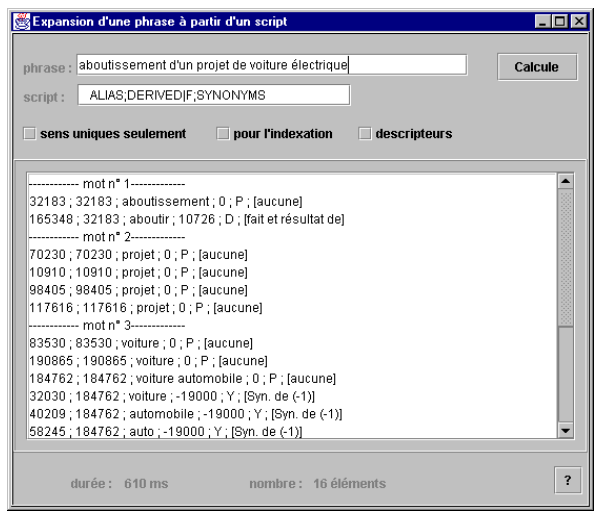

**Figure 2: Query expansion**

### 4.2 Word sense disambiguation and Term spotting

Lexical semantics provides an original approach for the term spotting task. Generally speaking, the main topics addressed by a document are expressed by ambiguous words. Most of the time, these words can be disambiguated from the context. If a document treats of *billiards*, the context of *billiards* is necessarily saturated by terms of larger topics like *games, competition*, *dexterity*... and terms in dependence with *billiard* like *ball, cue*, *cannon*...

Using this property, lexical topics are found by measuring the semantic proximity of each plain word of a text with the text itself. Terms that have the minimal semantic proximity are the best descriptors.

Note that this property may be used to verify the relevance of keywords manually given by a writer. An application may be the struggle to the spamming of search engine. To give an example of result of lexical summary, the algorithm applied to this paper provides in the 20 best words the terms : *lexicon, dictionary, semantic network, semantics, measures* and *disambiguation.* All these terms are highly relevant.

### 4.3 Emails sorting and answering

In this application, we have to classify a flow of documents according to a set of existing profiles. Most systems execute this task after a learning phase. A learning phase causes a problem because it needs a costly preliminary manual tagging of documents. It is then attractive to see if a complex lexicon could perform an accurate classification without any learning phase.

In our experiments the end-user must have to define profiles that correspond to his domains of interest. The formalism is very light: firstly, we define an identifier for each profile; secondly we define a definition of this profile (a set of relevant terms according to the domain). On the following examples, identifiers are given between parentheses and definitions are given after.

```
[guerre du Kosovo] guerre du Kosovo
[tabac et jeunesse] tabac et jeunesse
[alcoolisme et Bretagne] alcoolisme et Bretagne
[investissement immobilier en Ile-de-France] achat, vente et marché immobilier en Île-de-France
```

The definitions may be given in English with the exactly same result. The following text :

*Les loyers stagnent à Paris mais la baisse de la TVA sur les dépenses de réparation de l'habitat devrait soutenir le marché de l'ancien*

gives in term of semantic proximity:

```
[guerre du Kosovo]                       135
[tabac et jeunesse]                      140
[alcoolisme et Bretagne]                 129
[investissement immobilier en
Ile-de-France]                             9
```

We observe that differences between the mailboxes are very marked (the best score is the lowest one). Note that this approach may be used to help the classifying of web sites that is today entirely manually carry out.

## 5 Conclusion

In this paper, we have shown an efficient algorithm to semi-automatically acquire knowledge from a semantic network and a corpus. A set of basic services are also available through java APIs developed above the semantic network. We have shown that this set of elements offers a versatile toolbox for a large variety of NLP applications.